%% file: MemHand.tex
\documentclass[letterpaper, 10 pt, conference]{ieeeconf}
\IEEEoverridecommandlockouts    
\usepackage{hyperref}
\usepackage{makecell}
\usepackage{threeparttable}
\usepackage{caption}
\usepackage{soul}
\usepackage{color}
\usepackage{array}
\usepackage[table]{xcolor}
\usepackage{colortbl}
\usepackage{stfloats}
\usepackage{pifont} 
\renewcommand{\thetable}{\arabic{table}}
\makeatletter
\renewcommand{\fnum@table}{Table~\thetable}
\makeatother
\usepackage{etoolbox}
\hypersetup{
    colorlinks=true,
    linkcolor=magenta,
    filecolor=magenta,      
    urlcolor=magenta,
}

\input{stachnisslab-latex}


\usepackage{bm}
\usepackage{multirow}
\usepackage{rotating}  

\usepackage{multirow}
\usepackage{colortbl}
\definecolor{oursbg}{gray}{0.9}
\usepackage{tabularx}
\newcolumntype{Y}{>{\centering\arraybackslash}X}
\definecolor{secondbest}{RGB}{0,150,60}
\newcommand{\fst}[1]{\textbf{#1}}
\newcommand{\snd}[1]{\textcolor{secondbest}{#1}}
\newcommand{\na}{--}

\title{\LARGE \bf MemHand: Memory-Enhanced World-Space Hand Motion Reconstruction from Egocentric Videos}

\author{Junyi Ma$^{1\dagger}$, Zhangshuo Qi$^{2\dagger}$, Erhang Zhang$^{1}$, Mingjie Li$^{1}$, and Hesheng Wang$^{1*}$%
\thanks{$^{1}$Junyi Ma, Erhang Zhang, Mingjie Li, and Hesheng Wang  are with IRMV Lab, Shanghai Jiao Tong University, Shanghai 200240, China.}%
\thanks{$^{2}$Zhangshuo Qi is with School of Mechanical Engineering, Beijing Institute of Technology, Beijing, 100081, China.}%
\thanks{$^{\dagger}$Equal Contribution}
\thanks{$^{*}$Corresponding author email: wanghesheng@sjtu.edu.cn}%
}

\begin{document}

\makeatletter
\newtoks\IEEEthankstoks
\IEEEthankstoks{}
\let\IEEEorigfootnotetext\footnotetext
\renewcommand{\footnotetext}[2][\@empty]{%
  \def\@tempa{#1}%
  \ifx\@tempa\@empty
    \IEEEorigfootnotetext{#2}%
    \global\IEEEthankstoks\expandafter{\the\IEEEthankstoks
      \IEEEorigfootnotetext{#2}}%
  \else
    \IEEEorigfootnotetext[#1]{#2}%
    \global\IEEEthankstoks\expandafter{\the\IEEEthankstoks
      \IEEEorigfootnotetext[#1]{#2}}%
  \fi
}
\twocolumn[{%
  \renewcommand{\twocolumn}[1][]{#1}%
  \patchcmd{\@maketitle}{\vskip0.25in}{\vskip0pt}{}{}%
  \maketitle
  \centering
  \includegraphics[width=\textwidth]{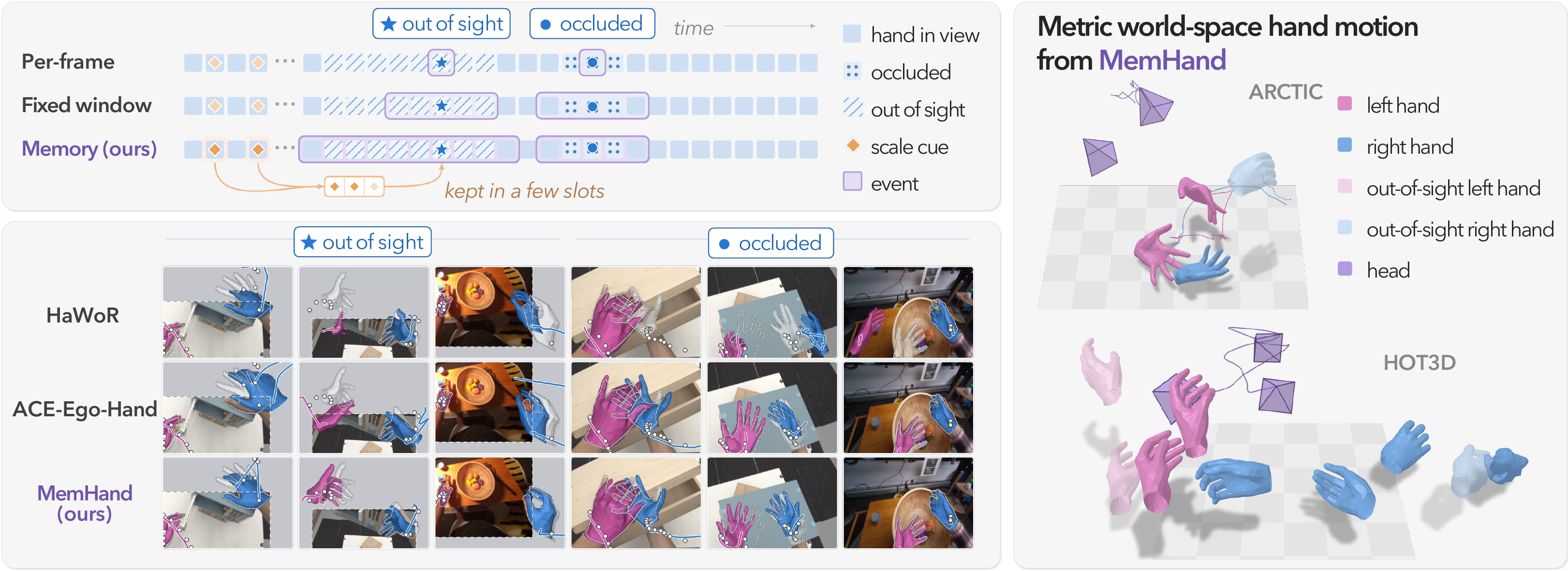}
  \captionof{figure}{\textbf{MemHand robustly reconstructs the metric world-space motion of both hands.} Compared to methods relying on single-frame or fixed-window inputs, MemHand leverages temporal context of matching length to process each event and maintains scale cues in a few memory slots. Qualitative comparisons against HaWoR and ACE-Ego-Hand showcase the benefits of the memory mechanism.}
  \label{fig:motivation}
  \vspace{1.2em}
}]
\let\footnotetext\IEEEorigfootnotetext
{\def\thefootnote{}\footnotesep 0.7\baselineskip \the\IEEEthankstoks}
\makeatother

\IEEEpeerreviewmaketitle
\thispagestyle{empty}
\pagestyle{empty}

\begin{abstract}

Recovering the continuous world-space motion of both hands from egocentric video is essential for applications such as human-to-robot policy transfer and augmented reality. Achieving this goes beyond reconstructing the hands from each individual frame, as the context that a frame needs is often spread over time, especially around occlusions or when a hand moves out of sight. Recent works have explored how to exploit temporal context, yet they mostly draw it from fixed-length windows. 
Such windows cannot guarantee that every frame has the context required to reconstruct the hands in world space with accurate poses and metric scale.
To address this limitation, we introduce MemHand, a memory-enhanced framework for world-space hand motion reconstruction. MemHand maintains two complementary forms of memory. The Short-Term Event Memory keeps the context of the well-observed hand around each occlusion or out-of-sight event and supplies it to frames with weak or missing visual evidence. In addition, since the metric scale holds over the whole recording, the Long-Term Scale Memory retains the most reliable observations across the video as scale anchors within a bounded state. 
Across benchmarks featuring severe occlusions and frequent out-of-sight intervals, MemHand achieves state-of-the-art accuracy and generalizes well to unseen domains. Benefiting from the synergy of both memories, it cuts W-MPJPE by 57\% on ARCTIC and 70\% on HOT3D relative to StableHand, delivering continuous and metrically accurate world-space hand poses and trajectories. We will release our code and models at the project page: \url{https://irmvlab.github.io/memhand.github.io}

\end{abstract}

\section{Introduction}
\label{sec:intro}

Egocentric video is emerging as a scalable way to collect demonstrations for robot learning~\cite{ma2026robot,grauman2022ego4d,hoque2025egodex}. The hands in these videos carry valuable motion knowledge and interaction patterns with the surroundings~\cite{kareer2025egomimic,ma2026egoloc,liu2025egozero,liu2022joint}, making hand reconstruction an important interface between human videos and robot policies~\cite{shaw2023videodex,qin2022dexmv,yang2025egovla,luo2025beingh0}. While single-frame hand reconstruction has advanced considerably in recent years, the target for a policy is typically the metric world-space motion of both hands over the video rather than per-frame estimates. Accordingly, world-space hand motion reconstruction has received growing attention~\cite{zhang2025hawor,yu2025dynhamr,zeng2026stablehand}.

Relative to single-frame hand reconstruction, an important problem in world-space motion estimation is temporal consistency. Across a video, the hand motion has to remain continuous even through challenging scenarios such as occlusions and out-of-sight intervals. To this end, existing video-based methods mostly introduce temporal context from a fixed-length window, using multi-view observations and motion cues over time to supplement the understanding of the hand's state on frames with limited visual evidence~\cite{zhang2025hawor,wang2026vidihand,liu2026aceegohand,moon2026rlhnd,zeng2026stablehand,ren2026infinihand}.

However, these methods implicitly assume that such a fixed window always captures sufficient temporal context. In practice, the context for any one hand is spread unevenly, and where it lies depends on the hand's state. For example, the position of an out-of-sight hand benefits from where it leaves and where it returns, and an occluded hand is best reconstructed from the context before and after the occlusion. Moreover, the metric scale of the trajectory is pinned down by a few moments that may occur anywhere in the video. 
Therefore, constrained by a predefined window size, they cannot reliably capture the temporal context that an accurate estimate at each frame actually requires, limiting further improvements in hand motion reconstruction accuracy.

To address this issue, we instead treat the effective use of temporal context as a memory problem. Because different kinds of context in world-space hand reconstruction have different lifetimes, we replace the fixed window with two complementary forms of memory.
Our Short-Term Event Memory segments the video into flexible-length events such as occlusions or out-of-sight gaps, each spanning from when the hand disappears to when it reappears, and thus encompassing well-observed hands at both ends. Because the temporal relevance of hand motion decays quickly after an event ends, the short-term memory stores dense visual evidence only for the current event duration, enabling whole-event decoding once the event closes. This ensures that a frame with less visual evidence can draw on the context within its event. The context for recording-level constants such as the metric scale matters indefinitely but arrives sparsely and with uneven reliability. Our Long-Term Scale Memory collects candidate observations, from both metric depth estimates and parallax formed by a static hand and a moving head, and retains the most precise ones for scale estimation. Through the complementary combination of the two memories, we achieve accurate world-space hand motion reconstruction. Our contributions are fourfold:

\begin{itemize}
  \item \textbf{A novel paradigm for hand motion reconstruction.} We present MemHand, a memory-enhanced framework that exploits temporal context to achieve accurate and robust world-space hand motion reconstruction.
  \item \textbf{Short-term event memory.} We propose a short-term memory that stores dense context across each occlusion or out-of-sight event, supplying frames that lack visual evidence with hand state cues beyond a fixed window.
  \item \textbf{Long-term scale memory.} A long-term memory mechanism is designed to preserve the sparse observations scattered across the whole video for accurate scale estimation.
  \item \textbf{Extensive experiments.} 
  Evaluations across multiple benchmarks show that MemHand outperforms state-of-the-art baselines, with ablations confirming the effectiveness of both memories in handling occlusions, out-of-sight intervals, and metric scale estimation.
\end{itemize}

\section{Related Work}
\label{sec:related_work}

\textbf{Hand reconstruction in camera space.}
Most monocular methods regress the parameters of the MANO hand model~\cite{romero2017mano} from a cropped image. HaMeR~\cite{pavlakos2024hamer} showed the effectiveness of scaling a ViT regressor, and WiLoR~\cite{potamias2025wilor} couples reconstruction with detection. To leverage temporal context, Deformer~\cite{fu2023deformer} and OmniHands~\cite{lin2024omnihands} fuse features within fixed-length windows of sampled frames. Recent approaches take this context from foundation models. ViDiHand~\cite{wang2026vidihand} and ACE-Ego-Hand~\cite{liu2026aceegohand} decode hands from Wan~\cite{wan2025wan}, and RLHND~\cite{moon2026rlhnd} from Cosmos~3~\cite{nvidia2026cosmos3}, but context beyond the fixed window is still hard to obtain~\cite{liu2026aceegohand}. In this work, we move beyond fixed windows, allowing each frame's context to follow the event it belongs to.

\noindent \textbf{Hand reconstruction in world space.}
Reconstruction in world space additionally estimates the motion of the camera and recovers the metric scale. Dyn-HaMR~\cite{yu2025dynhamr} optimizes camera pose and hand motion jointly. HaWoR~\cite{zhang2025hawor} estimates the camera pose by egocentric SLAM scaled with Metric3D~\cite{hu2024metric3dv2}. StableHand~\cite{zeng2026stablehand} takes the camera from a geometry foundation model. However, these methods refine per-frame hand estimates~\cite{pavlakos2024hamer,potamias2025wilor}, so occluded or out-of-sight hands often begin from a weak or missing prior. InfiniHand~\cite{ren2026infinihand} instead streams over sliding windows and carries LingBot-Map's memory~\cite{chen2026lingbotmap}, but this memory serves the camera rather than the hands, hands out of sight are not completed, and the scale relies on post-processing. In this work, we read the metric scale from our long-term memory of the most precise observations, and complete a hand out of sight to produce the continuous hand motion.

\noindent \textbf{Memory mechanisms for temporal context.}
Exploiting temporal context spread across a video has been actively studied in video generation and streaming 3D reconstruction. WorldMem~\cite{xiao2025worldmem} and Context-as-Memory~\cite{yu2025contextasmemory} keep past frames and retrieve them by camera pose or field-of-view overlap. StreamVGGT~\cite{zhuo2025streamvggt} keeps the cached tokens of past frames as implicit memory, and LingBot-Map~\cite{chen2026lingbotmap} divides its memory into anchor frames, a reference window, and a compressed trajectory. Hand reconstruction depends on such context just as much, but in a different form from the visual or geometric consistency these methods seek. The relevant context is largely determined by the hand state itself. In this work, we organize short-term and long-term memories around this property of hand reconstruction.

\begin{figure*}[ht]
  \centering
  \includegraphics[width=\linewidth]{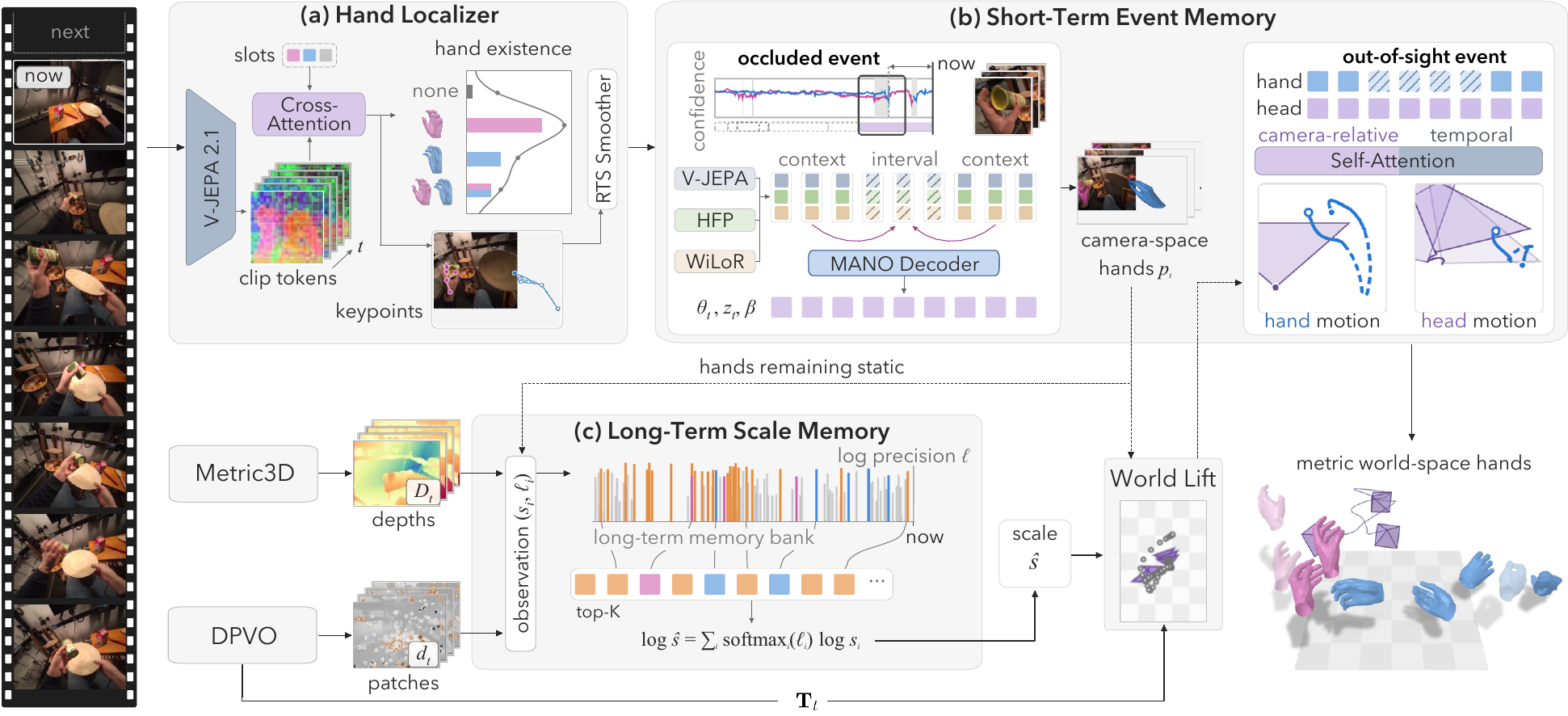}
  \caption{\textbf{System overview of MemHand.} The Hand Localizer (a) decides which hands are in each frame and tracks hand keypoints (Sec.~\ref{sec:localizer}). The Short-Term Event Memory (b) holds each event with context of its own length until the event closes. An occluded event is read in the camera frame from the WiLoR, V-JEPA and HFP features, and an out-of-sight event is completed in the world frame with the synergy between hand and head motion (Sec.~\ref{sec:event}). The Long-Term Scale Memory (c) keeps the most precise observations of the metric scale, from both metric depth and parallax of a static hand against head motion (Sec.~\ref{sec:scale}).}
  \label{fig:overview}
  \vspace{-8pt}
\end{figure*}

\section{Method}
\label{sec:method}


Given an egocentric video $\{I_t\}_{t=1}^{T}$, MemHand reconstructs both hands of the camera wearer as MANO~\cite{romero2017mano} meshes in metric world space, as illustrated in Fig.~\ref{fig:overview}. MemHand organizes temporal context at three ranges. Within fixed windows, the Hand Localizer decides which hands appear in each frame and tracks their keypoints from the features of a frozen V-JEPA~2.1 encoder~\cite{murlabadia2026vjepa21}. Beyond a single window, since an occlusion or out-of-sight gap may last longer, the Short-Term Event Memory retains context matching each such event's variable length to support the reconstruction of both occluded and out-of-sight hands. Spanning the whole recording, the Long-Term Scale Memory keeps the most precise observations of the metric scale within a bounded state. Connecting these three ranges, MemHand localizes in-image hands over fixed windows, reconstructs them in the camera frame with the short-term memory, and lifts them into metric world space along the camera trajectory scaled by the long-term memory. In this world space, each out-of-sight gap is completed from its event context, yielding continuous hand motion.

\subsection{Hand Localizer}
\label{sec:localizer}


Given the tokens $\mathbf{f}$ of a fixed window from the frozen V-JEPA encoder, the Hand Localizer first decides whether each hand $h\in\{\mathrm{L},\mathrm{R}\}$ appears in each frame $t$, denoted $e^{h}_{t}$. Whenever a hand is present, it then predicts for each keypoint $k$ an in-image indicator $v^{h,k}_{t}$, a 2D position $\mathbf{x}^{h,k}_{t}$, and an observation confidence $m^{h,k}_{t}$, as shown in Fig.~\ref{fig:overview} (a). Together, these outputs serve two purposes: (i) detecting and localizing in-image hands, a relatively simple task for which a fixed window suffices~\cite{liu2026aceegohand}, and (ii) identifying frames where visual evidence is degraded or missing, thus signaling when temporal context from our memory modules must be invoked to complement the fixed window.

\noindent \textbf{Hand existence.}
We formulate hand existence as identifying the states of the camera wearer's two hands, rather than localizing every hand as in~\cite{potamias2025wilor}. Following slot attention~\cite{locatello2020slotattention}, three slot queries representing the left hand, right hand, and background first attend to per-frame clip tokens $\mathbf{f}_t$ via soft assignment, preventing both hand slots from claiming the same hand. The left and right slots are then decoded into a distribution $\pi_t$ over $\{\varnothing,\mathrm{L},\mathrm{R},\mathrm{LR}\}$, denoting the states of none, left only, right only, and both. Finally, a hand is counted as present when $e^{\mathrm{L}}_{t}=\mathbf{1}\!\left[\pi_{t}(\mathrm{L})+\pi_{t}(\mathrm{LR})>0.5\right]$, with $e^{\mathrm{R}}_{t}$ defined by exchanging $\mathrm{L}$ for $\mathrm{R}$. 

\noindent \textbf{Keypoint localization.}
For each present hand, we then localize its 2D keypoints $\mathbf{x}^{h}_t$ from $\mathbf{f}$. We first apply a CenterNet-style decoder~\cite{zhou2019objects} to predict an image-space heatmap for each keypoint and read out its initial position. The probability mass around each peak of the heatmap defines the keypoint confidence $m^{h,k}_t$, which also identifies occluded events (Sec.~\ref{sec:event}). Since these heatmap estimates are reliable only within the image, an in-image head determines whether each keypoint lies inside, denoted $v^{h,k}_t$. For in-image keypoints, an auxiliary decoder predicts frame-to-frame displacements $\hat{\mathbf{u}}_t$ from $\mathbf{f}$. An RTS smoother~\cite{rauch1965rts} then fuses the initial positions, $\hat{\mathbf{u}}_t$, and variances derived from $m^{h,k}_t$ for temporal refinement. For out-of-image keypoints, we transform each keypoint's most recent in-image position into the current frame via the camera pose as a prior, from which a learned offset head recovers its position.

\subsection{Short-Term Event Memory}
\label{sec:event}


While the Hand Localizer's fixed window is mostly adequate for in-image hand localization, it does not always suffice for hand reconstruction, especially when a hand is occluded or moves out of sight. During such intervals, the contextual information needed to resolve the hand may lie beyond the window, as motion trends and interaction cues unfold over longer temporal spans. The Short-Term Event Memory therefore provides the second range of context, flexibly extending across fixed-window boundaries to capture each complete event, as shown in Fig.~\ref{fig:overview} (b).

\begin{figure}[t]
  \centering
  \includegraphics[width=\linewidth]{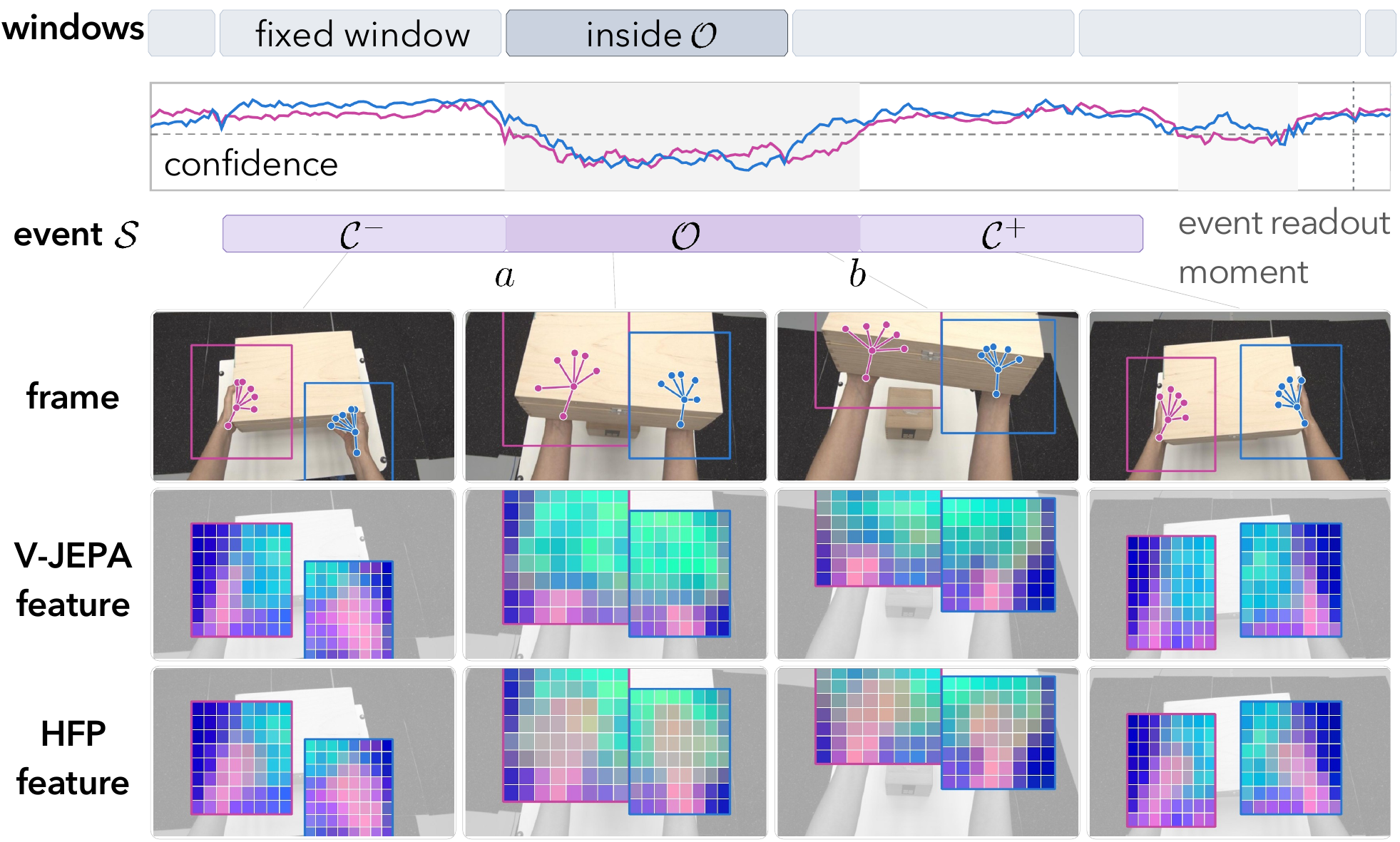}
  \caption{\textbf{An occluded event in the stream.} Top: Hand confidence drops below the threshold over an occlusion, leaving an entire fixed window inside the occlusion interval. MemHand pads the interval with visible leading and trailing contexts into an event. Bottom: V-JEPA features degrade under occlusion whereas HFP recovers features resembling those of visible hands.}
  \label{fig:event}
  \vspace{-12pt}
\end{figure}

\noindent \textbf{Event definition.}
We define two kinds of events, occluded and out of sight. In either kind, an event is an interval together with a \emph{leading} context and a \emph{trailing} context. For an occlusion, the interval $\mathcal{O}^{h}=[a,b)$ spans the frames where the mean of $m^{h,k}_t$ over a hand's in-image keypoints stays below a threshold. It rests on the prior that occlusion lowers the confidence of visual evidence that the fixed window cannot compensate for. This criterion requires neither labels nor a learned gate. As the length of an occlusion is unknown in advance, the short-term memory maintains a rolling buffer of the most recent $w_{\max}$ frames, so the leading context stays available. Once confidence recovers, we set $w=\operatorname{clip}(b-a,\,w_{\min},\,w_{\max})$, giving the leading context $\mathcal{C}^{-}=[a-w,a)$ and trailing context $\mathcal{C}^{+}=[b,b+w)$.
Consequently, we formulate the complete occluded event as $\mathcal{S}=\mathcal{C}^{-}\cup\mathcal{O}^{h}\cup\mathcal{C}^{+}$, as shown in Fig.~\ref{fig:event}. For an out-of-sight interval $\mathcal{U}^{h}$, we organize the leading and trailing context in essentially the same way as for an occluded event, thereby forming an out-of-sight event. After an event is resolved, the buffered evidence is released to keep the memory bounded.

\noindent \textbf{Decoding an occluded event.}
Before detailing the readout for occluded events, we define how our MANO Decoder decodes camera-space hands $p_{t}$ from a fixed window that is not assigned to any event. We first feed two input streams into the MANO Decoder, consisting of features from the frozen WiLoR ViT~\cite{potamias2025wilor} and V-JEPA, both sampled from the hand crop region. Following~\cite{potamias2025wilor}, we define a set of queries representing 6D rotations~\cite{zhou2019continuity}, wrist depth, and hand shape. These queries then perform cross-attention with the two feature streams. Subsequently, they are augmented with a reliability embedding encoded from the confidence $m_t$ and the keypoint in-image ratio. Temporal layers with RoPE-based frame-time embeddings~\cite{su2024roformer} then enable cross-frame attention among queries over the window. Finally, these tokens are decoded into 16 per-joint rotations $\theta_t$, wrist depth $z_t$, and per-window hand shape $\beta$. After aligning the MANO keypoints to $\mathbf{x}_{t}$ via a closed-form solution to compute the lateral position, we obtain the camera-space hands.

When decoding an occluded event, the MANO Decoder reads the full event $\mathcal{S}$ from the short-term memory rather than a fixed window, allowing sufficient information exchange across the occlusion. In addition to this extended context, we design a Hand Feature Prediction (HFP) module that serves as a third input stream to the MANO Decoder during occlusions. The motivation is that when the hand is occluded, the other two input streams predominantly capture features of the occluder, as shown in Fig.~\ref{fig:event}. However, ViDiHand~\cite{wang2026vidihand} and ACE-Ego-Hand~\cite{liu2026aceegohand} show that hands can still largely be decoded through occlusion from temporal video features. 
Inspired by them, we explicitly recover the occluded hand's representation in the V-JEPA feature space. Specifically, HFP is a transformer operating on the memorized crops $\mathcal{B}^{h}_{t}$ of both hands across all frames of $\mathcal{S}$, predicting a feature increment $\hat{\Delta}^{h}_{t}$ for each hand. Letting $\tilde{I}$ denote the reference video obtained by rendering the ground-truth MANO mesh onto the raw video $I$, $\hat{\Delta}^{h}_{t}$ represents the corresponding V-JEPA feature increment:
\begin{equation}
  \Delta^{h}_t = F_t(\tilde{I})[\mathcal{B}^{h}_{t}] - F_t(I)[\mathcal{B}^{h}_{t}] ,
  \label{eq:hfp}
\end{equation}
where $F_t(\cdot)[\mathcal{B}^{h}_{t}]$ is the feature of frame $t$ cropped at $\mathcal{B}^{h}_{t}$.

\noindent \textbf{Completing an out-of-sight event.}
An out-of-sight event differs from an occluded event in that its interval $\mathcal{U}^{h}$ lacks visual evidence of the hand. Meanwhile, a hand going out of sight may be driven by both head movement and hand motion, which are entangled in camera space. We therefore complete the memorized event in world space, where the two motions are naturally decoupled and hand-head synergy can compensate for the missing visual evidence in the out-of-sight interval. Specifically, in-image hands are first lifted into world space with scaled DPVO~\cite{teed2023dpvo} camera poses (Sec.~\ref{sec:scale}). Across $\mathcal{U}^{h}$, we initialize coarse hand poses from the lifted hands in the available leading and trailing context held by the short-term memory via interpolation or extrapolation. A completion module then refines these coarse poses into the completed hands $\hat{X}^{h}_t$. To exploit head movement, the module injects it into the self-attention layers. In every attention head, half of the channels are modulated by per-frame camera rotations~\cite{miyato2024gta} and the other half by timestamps~\cite{su2024roformer}, allowing attention to capture this synergy and improve out-of-sight completion accuracy.

In summary, the Short-Term Event Memory flexibly groups degraded intervals with their surrounding clear observations into self-contained events, retaining dense context only until each event is resolved. Across both occlusions and out-of-sight gaps, it thereby sustains continuous hand reconstruction beyond fixed-window limits.

\subsection{Long-Term Scale Memory}
\label{sec:scale}

While the Short-Term Event Memory supplies event-level context beyond the fixed window, the invariants that hold across a whole video, such as hand shape and metric scale, still need a third range of context. Benefiting from the first two ranges, our predicted hand shape is largely stable, whereas the scale from Metric3D~\cite{hu2024metric3dv2} alone lacks sufficient accuracy. The Long-Term Scale Memory therefore retains sparsely distributed anchors from both the reconstructed hands and monocular depth across the whole recording for accurate scale estimation, as shown in Fig.~\ref{fig:overview} (c).

\noindent \textbf{Observation modeling.}
The long-term memory records two forms of observations, namely depth observations and hand observations. Each observation $i$ maintains a value $s_i$ and a log precision $\ell_i$, where $s_i$ is a scale candidate derived from this observation, and $\ell_i$ indicates how much it should be trusted. A depth observation is taken at a keyframe where metric depth can be compared with visual odometry. Over the DPVO patches $\mathcal{P}_i$, the value is formulated as:
\begin{equation}
  \log s_i = \log \operatorname*{median}_{j\in\mathcal{P}_i}\bigl(D_j\, d_j\bigr),
  \label{eq:scene}
\end{equation}
where $D_j$ is the Metric3D depth at patch $j$, and $d_j$ is the DPVO inverse depth. For each observation, a lightweight network predicts its log precision $\ell_i$ from the per-patch depths, patch positions, and camera motion around the keyframe. Hand observations exploit the parallax between a static hand and head motion. Specifically, given the frequent motion of fingers, we represent the hand by its palm position, which is computed as the mean of the wrist and the four knuckles of the MANO keypoints. Over a short interval $\mathcal{W}_i$, let $\mathbf{q}_{t}$ be the camera-space palm position at frame $t$, and $\mathbf{T}_t$ the camera-to-world transform. An ideally static palm maintains a stable position, yielding the constraint:
\begin{equation}
  \sum_{t\in\mathcal{W}_i}\big\| \mathbf{T}_t\mathbf{q}_{t}-\overline{\mathbf{T}\mathbf{q}}\big\|^2=0 ,
  \label{eq:hand}
\end{equation}
where $\overline{\mathbf{T}\mathbf{q}}$ is the mean over $\mathcal{W}_i$. The translation scale of $\mathbf{T}_t$ is the only unknown, which we solve in closed form via least squares as $s_i$. Since a palm is rarely perfectly static over $\mathcal{W}_i$, we similarly predict its log precision $\ell_i$ from the palm trajectory, camera motion, and fitting residual to guide observation selection.

\noindent \textbf{Memory retention and scale readout.}
A long-term memory bank is introduced to maintain these observations throughout the recording. A new observation is retained when its $\ell_i$ is among the $K$ highest so far, and the memory of lowest precision leaves to make room, so that the bank maintains a bounded state for the whole recording. The scale is read from the bank in its current state as the precision-weighted mean of the values it holds:
\begin{equation}
  \log \hat{s} = \sum_{i} \operatorname{softmax}_i(\ell_i)\, \log s_i .
  \label{eq:recall}
\end{equation}

\subsection{Training Objectives}
\label{sec:training}

\noindent\textbf{Hand Localizer.}
The Hand Localizer is trained with the sum of five losses, namely a negative log-likelihood of hand existence, a binary cross-entropy loss for the in-image head, a cross-entropy against a Gaussian heatmap for the keypoint position decoder, a Huber loss on displacement prediction $\hat{\mathbf{u}}_t$, and an L1 loss on the learned offset head for out-of-image keypoints.

\noindent \textbf{MANO Decoder.}
Our MANO Decoder is supervised as in previous MANO regressors~\cite{pavlakos2024hamer,zhang2025hawor}. In addition, we apply L1 losses on the velocity and acceleration of the joints and of wrist depth $z_t$ to supervise the continuous motion throughout the event. HFP is trained separately from the MANO Decoder, with an L1 loss on the feature increment $\Delta^{h}_t$ at the occluded-event level.

\noindent \textbf{Completion module.}
Over each out-of-sight interval $\mathcal{U}^{h}$, the completion module minimizes the sum of per-joint Euclidean position and velocity errors between $\hat{X}^{h}_t$ and the ground truth, together with the second-order acceleration error of the hand centroid.

\noindent \textbf{Long-Term Scale Memory.}
The scale memory minimizes:
\begin{equation}
  \mathcal{L}_{\mathrm{scale}}=\Bigl|\log\frac{\hat s}{s^{*}}\Bigr|+\frac{\lambda_\ell}{N}\sum_{i}\mathcal{L}_{\mathrm{Lap}}\!\Bigl(\log\frac{s_i}{s_i^{*}},\,\ell_i\Bigr),
  \label{eq:scale}
\end{equation}
where $s^{*}$ is the scale of the similarity transform that aligns the camera trajectory to the ground truth. The second term averages over all $N$ collected observations the Laplace negative log-likelihood~\cite{kendall2017uncertainties} of each observation's own error, where $s_i^{*}$ is the value that observation $i$ would take without error. This calibrates the log precision $\ell_i$.

\section{Experiments}
\label{sec:experiments}

\subsection{Experimental Setup}
\label{sec:experimental_setup}

\noindent \textbf{Datasets.}
We train on ARCTIC~\cite{fan2023arctic} and HOT3D~\cite{banerjee2025hot3d}, holding HOI4D~\cite{liu2022hoi4d} out for zero-shot evaluation. We follow the training and test splits of ACE-Ego-Hand~\cite{liu2026aceegohand} on ARCTIC and HaWoR~\cite{zhang2025hawor} on HOT3D. For HOI4D, we also follow the 166-clip evaluation setting of ACE-Ego-Hand. As its clip list is unreleased, we sample 166 clips of our own.

\noindent \textbf{Implementation details.}
We keep the V-JEPA 2.1 ViT-g encoder~\cite{murlabadia2026vjepa21} and the WiLoR ViT~\cite{potamias2025wilor} frozen during training. The encoder reads $384\times384$ frames in 64-frame windows. The context on either side of an event spans $w_{\min}=16$ to $w_{\max}=64$. The long-term memory bank keeps $K=32$ observations. Following the order of the pipeline, we train the modules sequentially, freezing previously trained ones at each stage. All modules use AdamW~\cite{loshchilov2019adamw} with gradient clipping at 1.0 and a linear learning rate warm-up followed by cosine decay. The Hand Localizer, HFP, the MANO Decoder, and the completion module are trained for 6, 12, 20, and 8 epochs, respectively, with a peak learning rate of $3\times10^{-4}$. The Long-Term Scale Memory is trained for 150 epochs with a learning rate of $10^{-3}$.

\begin{table*}[t]
  \caption{Camera-space hand reconstruction on ARCTIC, HOT3D, and HOI4D. Best and second-best results are in bold black and green.}
  \label{tab:camera}
  \centering
  \footnotesize
  \setlength{\tabcolsep}{2pt}
  \setlength{\aboverulesep}{0pt}
  \setlength{\belowrulesep}{0pt}
  \renewcommand{\arraystretch}{1.1}
  \def\hs{\rule[-0.86ex]{0pt}{3.2ex}}%
  \def\hc#1{\raisebox{1.6ex}[0pt][0pt]{#1}}%
  \begin{tabularx}{\textwidth}{c|l|YccYYYY|YcY}
    \toprule
    \hs & & \multicolumn{7}{c|}{In sight} & \multicolumn{3}{c}{Out of sight} \\
    \cline{3-12}
    \hs\hc{Dataset} & \multicolumn{1}{c|}{\hc{Method}} & FAcc$\uparrow$ & MPJPE-p$\downarrow$ & PA-MPJPE-p$\downarrow$ & EPE-p$\downarrow$ & GO-p$\downarrow$ & CT-p$\downarrow$ & Jitter$\downarrow$ & Cov.$\uparrow$ & MPJPE-p$\downarrow$ & OW-p$\downarrow$ \\
    \midrule
    \noalign{\global\def\arraystretch{0.95}}
    \multirow{6}{*}{ARCTIC}
    \rule[-1.0ex]{0pt}{3.2ex}& HaMeR~\cite{pavlakos2024hamer} & 0.875 & 29.20 & 14.60 & 65.29 & 24.91 & 0.095 & 18.28 & 0.177 & 151.53 & 823.29 \\
    & WiLoR~\cite{potamias2025wilor} & 0.919 & 22.01 & 11.87 & 71.53 & 17.36 & 0.075 & 24.09 & 0.004 & 162.33 & 875.78 \\
    & Dyn-HaMR~\cite{yu2025dynhamr} & 0.947 & 23.13 & 11.32 & 31.59 & 18.94 & 0.067 & 13.46 & 0.515 & 134.84 & 709.65 \\
    & HaWoR~\cite{zhang2025hawor} & 0.943 & 32.85 & 11.14 & 30.33 & 23.93 & 0.058 & 15.08 & 0.983 & 103.24 & 465.74 \\
    & ACE-Ego-Hand~\cite{liu2026aceegohand} & \fst{1.000} & \snd{15.26} & \snd{7.47} & \snd{9.18} & \snd{11.81} & \fst{0.021} & \fst{2.70} & \fst{1.000} & \snd{37.78} & \snd{119.32} \\
    \rowcolor{oursbg}\cellcolor{white}\rule[-1.0ex]{0pt}{3.2ex}& MemHand (ours) & \fst{1.000} & \fst{14.07} & \fst{6.97} & \fst{8.72} & \fst{11.14} & \fst{0.021} & \snd{3.29} & \fst{1.000} & \fst{36.70} & \fst{103.76} \\
    \midrule
    \multirow{6}{*}{HOT3D}
    \rule[-1.0ex]{0pt}{3.2ex}& HaMeR~\cite{pavlakos2024hamer} & 0.701 & 66.69 & 23.25 & 62.48 & 48.12 & 0.100 & 19.03 & 0.420 & 135.06 & 673.31 \\
    & WiLoR~\cite{potamias2025wilor} & 0.808 & 27.66 & 20.57 & 71.97 & 23.40 & 0.096 & 12.12 & 0.031 & 146.60 & 645.11 \\
    & Dyn-HaMR$^{*}$~\cite{yu2025dynhamr} & 0.759 & 29.06 & 14.28 & 17.74 & 20.93 & 0.074 & 14.10 & \snd{0.995} & 78.55 & 367.11 \\
    & HaWoR$^{\ddagger}$~\cite{zhang2025hawor} & \snd{0.874} & \snd{16.89} & \snd{9.30} & \snd{16.97} & \snd{11.44} & \snd{0.040} & \snd{11.89} & 0.977 & \snd{71.77} & \snd{318.91} \\
    & \textcolor{gray}{ACE-Ego-Hand$^{\mathsection}$~\cite{liu2026aceegohand}} & \textcolor{gray}{0.991} & \textcolor{gray}{8.61} & \textcolor{gray}{4.39} & \textcolor{gray}{4.44} & \textcolor{gray}{4.99} & \textcolor{gray}{0.023} & \textcolor{gray}{3.22} & \textcolor{gray}{0.998} & \textcolor{gray}{20.94} & \textcolor{gray}{85.93} \\
    \rowcolor{oursbg}\cellcolor{white}\rule[-1.0ex]{0pt}{3.2ex}& MemHand (ours) & \fst{0.962} & \fst{7.84} & \fst{4.58} & \fst{4.05} & \fst{4.97} & \fst{0.012} & \fst{3.61} & \fst{1.000} & \fst{36.45} & \fst{100.80} \\
    \midrule
    \multirow{6}{*}{HOI4D}
    \rule[-1.0ex]{0pt}{3.2ex}& HaMeR~\cite{pavlakos2024hamer} & 0.726 & 49.34 & 24.09 & 95.07 & 38.34 & 0.208 & 23.51 & \na & \na & \na \\
    & WiLoR~\cite{potamias2025wilor} & 0.961 & 37.38 & 14.69 & 37.21 & 27.51 & 0.143 & 38.00 & \na & \na & \na \\
    & Dyn-HaMR$^{*}$~\cite{yu2025dynhamr} & 0.965 & 32.37 & 15.60 & 30.36 & 25.44 & 0.151 & 14.75 & \na & \na & \na \\
    & HaWoR~\cite{zhang2025hawor} & 0.942 & 35.04 & 13.00 & 21.74 & 25.16 & 0.058 & 21.90 & \na & \na & \na \\
    & ACE-Ego-Hand~\cite{liu2026aceegohand} & \snd{0.972} & \snd{26.14} & \snd{12.37} & \snd{19.65} & \snd{20.76} & \snd{0.052} & \fst{2.26} & \na & \na & \na \\
    \rowcolor{oursbg}\cellcolor{white}\rule[-1.0ex]{0pt}{3.2ex}& MemHand (ours) & \fst{0.995} & \fst{24.54} & \fst{10.53} & \fst{11.30} & \fst{18.46} & \fst{0.045} & \snd{3.55} & \na & \na & \na \\
    \bottomrule
  \end{tabularx}
  \par\vspace{2pt}
  \begin{minipage}{\textwidth}\scriptsize
    $^{\mathsection}$This method's training set overlaps with our test set, and its specific data splits are not publicly available, limiting direct comparability. The results are for reference only and not ranked.
    $^{*}$The sequences that failed in optimization are left out.
    $^{\ddagger}$The out-of-sight hands with non-finite wrist position are left out.
  \end{minipage}
  \par\vspace{0pt}
\end{table*}

\begin{table*}[t]
  \caption{World-space hand motion reconstruction on ARCTIC, HOT3D and zero-shot HOI4D. PA, W, and WA denote PA-MPJPE, W-MPJPE, and WA-MPJPE. Best and second-best results are in bold black and green.}
  \label{tab:world}
  \centering
  \footnotesize
  \setlength{\tabcolsep}{2pt}
  \setlength{\aboverulesep}{0pt}
  \setlength{\belowrulesep}{0pt}
  \renewcommand{\arraystretch}{1.1}
  \def\hs{\rule[-0.86ex]{0pt}{3.2ex}}%
  \def\hc#1{\raisebox{1.6ex}[0pt][0pt]{#1}}%
  \begin{tabularx}{\textwidth}{c|l|Y|YYY|YYY|YYY}
    \toprule
    \hs & & & \multicolumn{3}{c|}{Matched} & \multicolumn{3}{c|}{Out of sight} & \multicolumn{3}{c}{All} \\
    \cline{4-12}
    \hs\hc{Dataset} & \multicolumn{1}{c|}{\hc{Method}} & \hc{Recall$\uparrow$} & PA$\downarrow$ & W$\downarrow$ & WA$\downarrow$ & PA$\downarrow$ & W$\downarrow$ & WA$\downarrow$ & PA$\downarrow$ & W$\downarrow$ & WA$\downarrow$ \\
    \midrule
    \noalign{\global\def\arraystretch{0.95}}
    \multirow{5}{*}{ARCTIC}
    \rule[-1.0ex]{0pt}{3.2ex}& WiLoR-SLAM~\cite{potamias2025wilor,teed2021droid,hu2024metric3dv2} & 0.929 & 7.16 & 62.61 & 45.58 & 15.48 & \snd{591.28} & 294.25 & \snd{8.44} & \snd{130.55} & \snd{133.51} \\
    & Dyn-HaMR~\cite{yu2025dynhamr} & \snd{0.979} & 10.16 & 106.35 & 62.13 & 15.49 & 750.17 & 310.46 & 10.91 & 184.12 & 160.71 \\
    & HaWoR~\cite{zhang2025hawor} & 0.970 & 8.82 & 83.56 & 43.90 & \snd{14.37} & 650.39 & \snd{290.85} & 9.81 & 157.21 & 144.97 \\
    & InfiniHand$^{\dagger}$~\cite{ren2026infinihand} & \na & \snd{7.07} & \snd{59.21} & \snd{43.59} & \na & \na & \na & \na & \na & \na \\
    \rowcolor{oursbg}\cellcolor{white}\rule[-1.0ex]{0pt}{3.2ex}& MemHand (ours) & \fst{1.000} & \fst{6.97} & \fst{32.47} & \fst{22.84} & \fst{9.31} & \fst{111.47} & \fst{73.88} & \fst{7.23} & \fst{41.46} & \fst{33.29} \\
    \midrule
    \multirow{4}{*}{HOT3D}
    \rule[-1.0ex]{0pt}{3.2ex}& WiLoR-SLAM~\cite{potamias2025wilor,teed2021droid,hu2024metric3dv2} & 0.886 & 6.85 & 61.27 & 41.23 & 12.60 & 331.53 & 236.32 & 8.00 & 106.81 & 100.03 \\
    & Dyn-HaMR$^{*}$~\cite{yu2025dynhamr} & 0.943 & 11.65 & 183.50 & 121.34 & 15.99 & 411.37 & 240.76 & 12.50 & 225.92 & 169.09 \\
    & HaWoR$^{\ddagger}$~\cite{zhang2025hawor} & \snd{0.961} & \snd{6.55} & \snd{46.25} & \snd{35.89} & \snd{10.98} & \snd{316.57} & \snd{235.60} & \snd{7.34} & \snd{89.81} & \snd{95.75} \\
    \rowcolor{oursbg}\cellcolor{white}\rule[-1.0ex]{0pt}{3.2ex}& MemHand (ours) & \fst{0.999} & \fst{4.57} & \fst{24.84} & \fst{16.48} & \fst{7.13} & \fst{105.18} & \fst{93.21} & \fst{4.91} & \fst{35.51} & \fst{29.36} \\
    \midrule
    \multirow{4}{*}{HOI4D}
    \rule[-1.0ex]{0pt}{3.2ex}& WiLoR-SLAM~\cite{potamias2025wilor,teed2021droid,hu2024metric3dv2} & 0.967 & \snd{10.42} & 134.81 & 61.22 & \na & \na & \na & \snd{10.47} & 136.77 & 63.86 \\
    & Dyn-HaMR$^{*}$~\cite{yu2025dynhamr} & 0.972 & 12.10 & 307.65 & 99.34 & \na & \na & \na & 12.19 & 308.31 & 102.28 \\
    & HaWoR~\cite{zhang2025hawor} & \snd{0.983} & 11.22 & \snd{102.38} & \snd{52.21} & \na & \na & \na & 11.23 & \snd{103.07} & \snd{52.98} \\
    \rowcolor{oursbg}\cellcolor{white}\rule[-1.0ex]{0pt}{3.2ex}& MemHand (ours) & \fst{0.996} & \fst{10.34} & \fst{87.65} & \fst{38.13} & \na & \na & \na & \fst{10.35} & \fst{88.33} & \fst{38.91} \\
    \bottomrule
  \end{tabularx}
  \par\vspace{2pt}
  \begin{minipage}{\textwidth}\scriptsize
    $^{\dagger}$The numbers are quoted from its paper, which is produced on the same ARCTIC test set.
    $^{*}$The sequences that failed in optimization are left out.
    $^{\ddagger}$The out-of-sight hands with non-finite wrist position are left out.
  \end{minipage}
  \par\vspace{-6pt}
\end{table*}

\begin{figure*}[t]
  \centering
  \includegraphics[width=\linewidth]{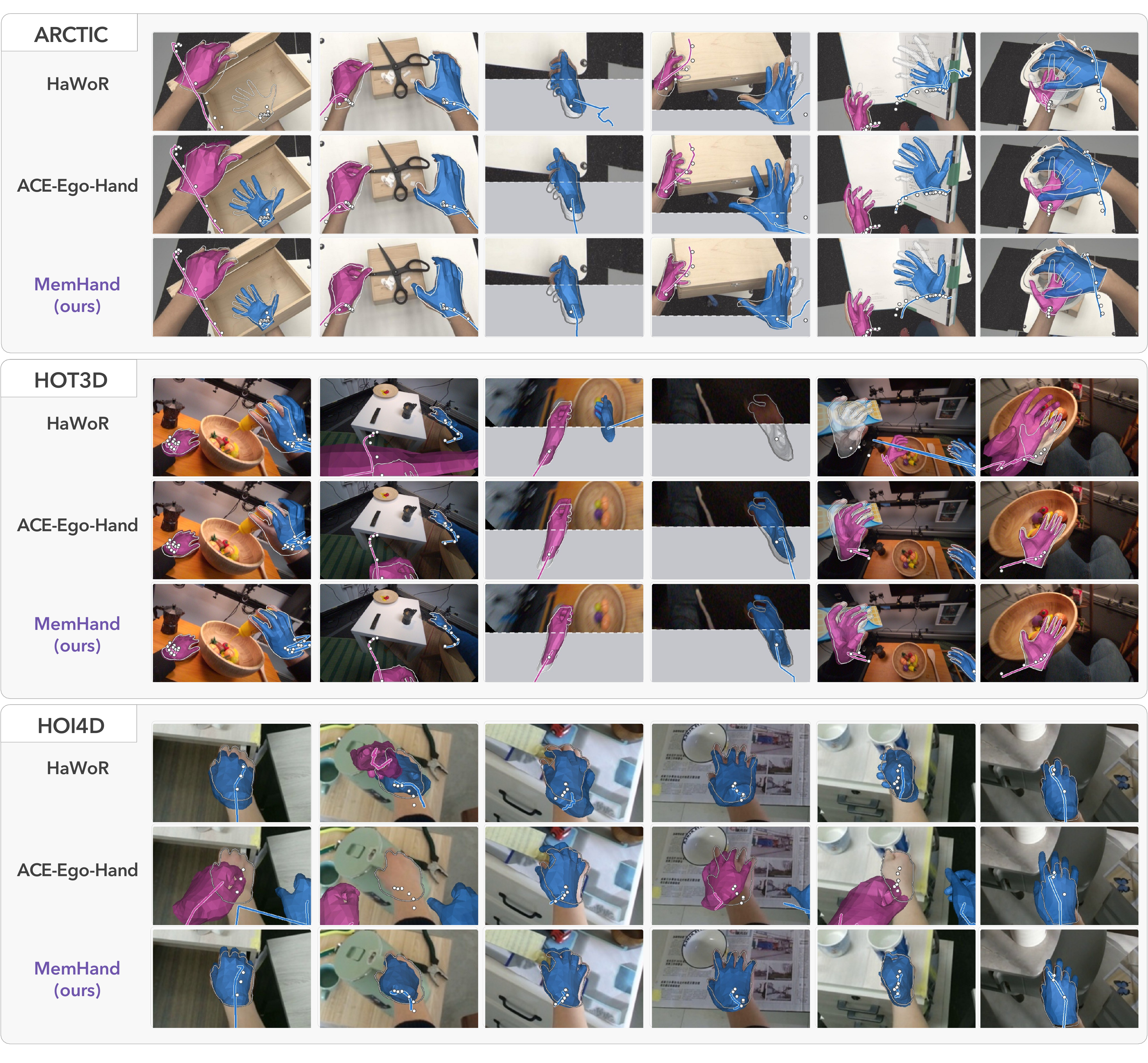}
  \caption{\textbf{Camera-space comparisons on ARCTIC, HOT3D, and zero-shot HOI4D.} Each column shows a frame where all the hands are in sight, or where a hand is out of sight or occluded. The ground-truth hands are outlined in white and filled in white where they are occluded or out of sight. For an out-of-sight or occluded hand, the ground-truth and predicted wrist trajectories are drawn over the whole event, and out-of-sight hands are drawn outside the image border.}
  \label{fig:qual_cam}
\end{figure*}

\begin{figure*}[t]
  \centering
  \includegraphics[width=\linewidth]{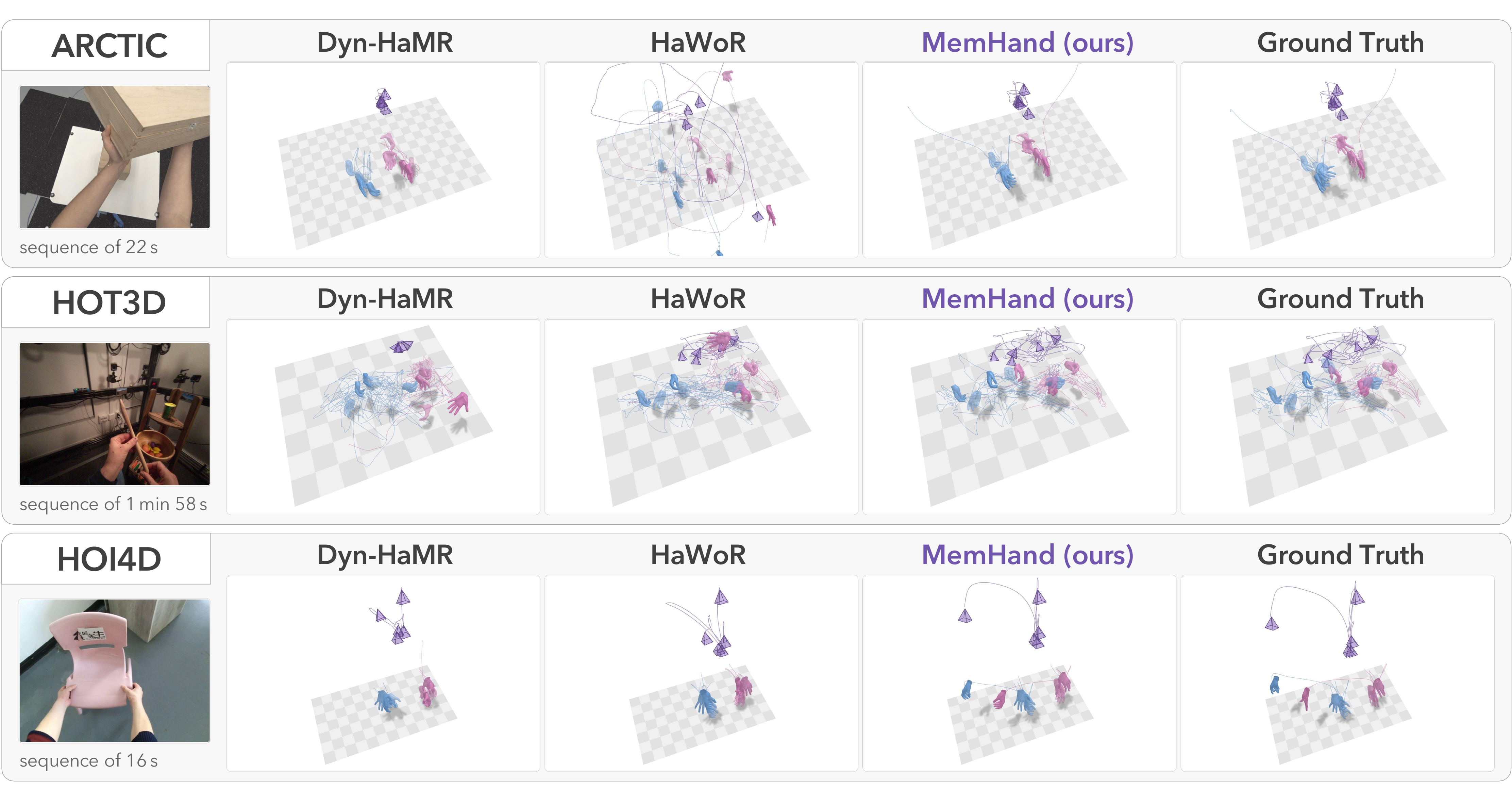}
  \caption{\textbf{World-space comparisons over whole sequences of ARCTIC, HOT3D, and HOI4D.} Each row shows the hand and head paths over a whole test sequence, with the hands and head poses at five evenly spaced instants. The image in the middle of the sequence is shown for reference.}
  \label{fig:qual_whole}
\end{figure*}

\subsection{Camera-Space Hand Reconstruction}
\label{sec:camera}

We first demonstrate that our combined short-term and long-term memory yields accurate camera-space hand reconstruction, the widely used evaluation setting in the field. The baselines and our MemHand are scored under the ACE-Ego-Hand~\cite{liu2026aceegohand} protocol. Notably, ACE-Ego-Hand evaluates out-of-sight hands only via a wrist-aligned error, neglecting their spatial placement. Alongside wrist-aligned MPJPE-p, we therefore introduce two out-of-sight metrics. Cov. is the ratio of completed missing frames, and OW-p is the mean Euclidean distance of non-aligned joints. The uncompleted out-of-sight hands are penalized in MPJPE-p and OW-p using a mean-shaped hand with identity rotation.

For ARCTIC, we share the same test set with ACE-Ego-Hand. Accordingly, we quote the results of HaMeR~\cite{pavlakos2024hamer}, WiLoR~\cite{potamias2025wilor}, and ACE-Ego-Hand from~\cite{liu2026aceegohand} directly, re-running them only to reproduce out-of-sight metrics. For Dyn-HaMR~\cite{yu2025dynhamr} and HaWoR~\cite{zhang2025hawor}, we re-evaluate both methods via camera-space reprojection of their world-space outputs, thereby assessing their full workflows. For HOT3D and HOI4D, we re-run the baselines on our own test sets for a fair comparison. On HOT3D, ACE-Ego-Hand uses a smaller test set than our HaWoR split, so its training set necessarily includes our test sequences. However, since its exact data splits are not publicly available, we cannot rigorously exclude this effect, so the HOT3D results are provided for reference only. Out-of-sight metrics are omitted on HOI4D because it annotates visible hands only.

As shown in Table~\ref{tab:camera}, in like-for-like comparisons, MemHand outperforms the baselines in nearly all metrics, with only ACE-Ego-Hand edging ahead on jitter suppression. Even on HOT3D, where ACE-Ego-Hand serves as a favorable reference, its clear lead is confined to the out-of-sight metrics, where predictions lack visual anchoring and are therefore much more prone to memorizing seen trajectories. Notably, our advantage is more evident on zero-shot HOI4D, underscoring the cross-domain generalization capability. More qualitative comparisons are shown in Fig.~\ref{fig:qual_cam}.

\subsection{World-Space Hand Reconstruction}
\label{sec:world}
We further validate that MemHand estimates accurate hand motion in world space. We follow the settings of InfiniHand~\cite{ren2026infinihand}, which evaluates the full video allowing error accumulation without per-window resets using W-MPJPE, WA-MPJPE, and PA-MPJPE. Since InfiniHand scores only the in-image ground-truth hands matched by each method and omits out-of-sight ones, we report these metrics on the matched hands, out-of-sight hands, and all ground-truth hands. Moreover, we report Recall, the share of in-image ground-truth hands matched by a method's output. As in InfiniHand, we evaluate WiLoR-SLAM~\cite{potamias2025wilor,teed2021droid,hu2024metric3dv2}, HaWoR~\cite{zhang2025hawor}, and Dyn-HaMR~\cite{yu2025dynhamr} locally as baselines. Besides, we directly quote InfiniHand's published results on ARCTIC, as it is evaluated on the same test split. Since HOI4D lacks out-of-sight annotations, the out-of-sight columns are omitted. As shown in Table~\ref{tab:world}, MemHand outperforms all baselines across all metrics. On matched hands, its PA-MPJPE stays relatively close to the baselines while W-MPJPE improves substantially, showing that our gain lies mainly in world-space hand placement. Notably, our clear margin persists on zero-shot HOI4D, demonstrating strong generalization in world-space hand motion estimation. Qualitative comparisons over whole test sequences are shown in Fig.~\ref{fig:qual_whole}, where all the trajectories are anchored at the first frame without further alignment.

Moreover, we evaluate MemHand on the benchmark of StableHand~\cite{zeng2026stablehand}, a SOTA world-space hand reconstruction method. Since the official release provides an enhanced version that outperforms the results reported in its paper, we use this version for our comparison. We follow its evaluation protocol, except on HOT3D where we retain only the test clips unseen by both methods for a fair comparison. As shown in Table~\ref{tab:stablehand}, MemHand outperforms StableHand on most metrics. The gain is most pronounced in W-MPJPE, which measures world-space hand placement accuracy over the entire clip. Specifically, MemHand reduces W-MPJPE by 57\% on ARCTIC and 70\% on HOT3D. Notably, StableHand uses privileged ground-truth hand shape in its released evaluation, whereas MemHand relies on its own predicted shape, further underscoring the effectiveness of our method.

\begin{table}[t]
  \caption{World-space hand motion reconstruction on the StableHand benchmark.}
  \label{tab:stablehand}
  \centering
  \footnotesize
  \setlength{\tabcolsep}{2pt}
  \setlength{\aboverulesep}{0pt}
  \setlength{\belowrulesep}{0pt}
  \renewcommand{\arraystretch}{1.1}
  \begin{tabularx}{\columnwidth}{c|l|YYYYc}
    \toprule
    \rule[-1.0ex]{0pt}{3.2ex}Dataset & \multicolumn{1}{c|}{Method} & PA$\downarrow$ & W$\downarrow$ & WA$\downarrow$ & Acc$\downarrow$ & MRRPE$\downarrow$ \\
    \midrule
    \noalign{\global\def\arraystretch{0.95}}
    \rule[-1.0ex]{0pt}{3.2ex}& StableHand~\cite{zeng2026stablehand} & 7.56 & 91.25 & 29.25 & \fst{3.31} & 88.56 \\
    \rowcolor{oursbg}\cellcolor{white}\rule[-1.0ex]{0pt}{3.2ex}\multirow{-2}{*}{ARCTIC} & MemHand (ours) & \fst{5.67} & \fst{39.05} & \fst{13.71} & 3.54 & \fst{35.27} \\
    \midrule
    \rule[-1.0ex]{0pt}{3.2ex}& StableHand~\cite{zeng2026stablehand} & 3.40 & 46.76 & 18.71 & 2.91 & 47.36 \\
    \rowcolor{oursbg}\cellcolor{white}\rule[-1.0ex]{0pt}{3.2ex}\multirow{-2}{*}{HOT3D} & MemHand (ours) & \fst{3.19} & \fst{14.20} & \fst{6.02} & \fst{2.42} & \fst{15.54} \\
    \bottomrule
  \end{tabularx}
  \par\vspace{0pt}
\end{table}

\subsection{Ablation Studies}
\label{sec:ablation}

\noindent \textbf{Memory mechanisms.} We first ablate the benefits of our memory mechanism, with results shown in Table~\ref{tab:ablation}. Following ACE-Ego-Hand, we adopt a reduced training recipe where all ablation variants are only trained on ARCTIC for half the epochs of the main pipeline. We first reconstruct the entire pipeline based on single-frame DINOv3 ViT-H+~\cite{simeoni2025dinov3}, denoted as (a). (b) allows DINOv3 feature fusion within a fixed window and outperforms (a). (c) replaces DINOv3 with V-JEPA over the same fixed window, achieving further gains over (b), which justifies our backbone choice. (d) incorporates short-term occluded event memory and yields additional improvements over (c), demonstrating the effectiveness of HFP features and the extended temporal context. (e) further adds short-term memory for out-of-sight events, substantially improving predictions for out-of-sight hands. The gains of (d) and (e) together validate the necessity of our short-term memory. Finally, (f) adds long-term scale memory, which further improves world-space motion accuracy compared to the scale estimated by Metric3D in (e). 

\begin{table}[t]
  \caption{Ablation of the memories. MPJPE-p and OW-p are the in-sight and out-of-sight errors of Table~\ref{tab:camera}, and W is the world-space error over all the frames of Table~\ref{tab:world}. Best and second-best results are in bold black and green.}
  \label{tab:ablation}
  \centering
  \footnotesize
  \setlength{\tabcolsep}{2pt}
  \setlength{\aboverulesep}{0pt}
  \setlength{\belowrulesep}{0pt}
  \renewcommand{\arraystretch}{1.1}
  \def\hs{\rule[-0.86ex]{0pt}{3.2ex}}%
  \def\hc#1{\raisebox{1.6ex}[0pt][0pt]{#1}}%
    \def\gr#1{\textcolor{black!55}{#1}}%
  \begin{tabularx}{\columnwidth}{c|l|c|ccc|cYY}
    \toprule
    \hs & & Fixed & \multicolumn{3}{c|}{Memory} & & & \\
    \cline{4-6}
    \hs & \multicolumn{1}{c|}{\hc{Backbone}} & window & Occ. & Out & Scale & \hc{MPJPE-p$\downarrow$} & \hc{OW-p$\downarrow$} & \hc{W$\downarrow$} \\
    \midrule
    \noalign{\global\def\arraystretch{0.95}}
    \rule[-1.0ex]{0pt}{3.2ex}(a) & DINOv3 &  &  &  &  & 16.66 & 259.66 & 76.87 \\
    (b) & DINOv3 & \checkmark &  &  &  & 15.95 & 251.50 & 73.07 \\
    (c) & V-JEPA 2.1 & \checkmark &  &  &  & 15.56 & 225.62 & 69.65 \\
    \midrule
    \rule[-1.0ex]{0pt}{3.2ex}(d) & V-JEPA 2.1 & \checkmark & \checkmark &  &  & \fst{14.88} & 218.74 & 68.55 \\
    (e) & V-JEPA 2.1 & \checkmark & \checkmark & \checkmark &  & \fst{14.88} & \snd{132.71} & \snd{44.79} \\
    \rowcolor{oursbg}\rule[-0.86ex]{0pt}{3.2ex}(f) & V-JEPA 2.1 & \checkmark & \checkmark & \checkmark & \checkmark & \fst{14.88} & \fst{131.07} & \fst{42.69} \\ \hline
    \rule[-0.86ex]{0pt}{3.2ex}\gr{(g)} & \gr{V-JEPA 2.1} & \gr{\checkmark} & \gr{\checkmark} & \gr{\checkmark} & \gr{GT} & \gr{14.88} & \gr{130.79} & \gr{41.77} \\
    \bottomrule
  \end{tabularx}
  \par\vspace{0pt}
\end{table}

To further validate our long-term memory, we include reference row (g) with the ground-truth scale aligned from the camera trajectory. As scale error accumulates with camera translation, the subtle tabletop head motion of ARCTIC leaves only a 3.0\,mm margin in W even with the ground-truth scale, of which our memory already captures ${\sim}70\%$. On HOT3D and zero-shot HOI4D, where camera translation is larger, the gain over Metric3D-based scale is much clearer, reducing W from 40.1 to 35.5\,mm and from 99.2 to 88.3\,mm, against 32.8 and 76.2\,mm with the ground-truth scale. Moreover, as more observations accumulate over time, the median scale error of our memory on HOT3D sequences longer than 60\,s steadily drops from 9.2\% at 5\,s to 3.4\% at 60\,s and 2.9\% over the full recording, demonstrating its progressive scale refinement.

\noindent \textbf{Input features for MANO Decoder.} We further ablate the input features of the MANO Decoder, namely V-JEPA, WiLoR ViT, and HFP. As shown in Table~\ref{tab:reader}, the spatiotemporal features of V-JEPA contribute significantly to wrist localization, whereas the texture-rich WiLoR ViT features benefit mostly the estimation of hand shape and pose. The combination of WiLoR and V-JEPA improves all metrics. Moreover, adding HFP yields the best overall results, although it is active only on occluded events. Furthermore, comparing the V-JEPA+WiLoR variant with (c) in Table~\ref{tab:ablation} isolates the MPJPE-p gain from longer temporal context on occluded events alone without HFP.

\begin{table}[t]
  \caption{Ablation of the input features of the MANO Decoder. The metrics are from Table~\ref{tab:camera} (CT-p reported in mm). Best and second-best results are in bold black and green.}
  \label{tab:reader}
  \centering
  \footnotesize
  \setlength{\tabcolsep}{2pt}
  \setlength{\aboverulesep}{0pt}
  \setlength{\belowrulesep}{0pt}
  \renewcommand{\arraystretch}{1.1}
  \def\hs{\rule[-0.86ex]{0pt}{3.2ex}}%
  \def\hc#1{\raisebox{1.6ex}[0pt][0pt]{#1}}%
  \begin{tabularx}{\columnwidth}{*{3}{>{\hspace{1pt}}c<{\hspace{1pt}}}|YYY}
    \toprule
    \multicolumn{3}{c|}{\hs Input features} & & & \\
    \cline{1-3}
    \hs V-JEPA & WiLoR & HFP & \hc{MPJPE-p$\downarrow$} & \hc{PA-MPJPE-p$\downarrow$} & \hc{CT-p$\downarrow$} \\
    \midrule
    \noalign{\global\def\arraystretch{0.95}}
    \rule[-1.0ex]{0pt}{3.2ex}\checkmark &  &  & 17.43 & 8.97 & 24.7 \\
     & \checkmark &  & 15.60 & 8.37 & 26.7 \\
    \checkmark & \checkmark &  & \snd{14.99} & \snd{7.80} & \snd{23.5} \\
    \rowcolor{oursbg}\rule[-0.86ex]{0pt}{3.2ex}\checkmark & \checkmark & \checkmark & \fst{14.88} & \fst{7.78} & \fst{22.9} \\
    \bottomrule
  \end{tabularx}
  \par\vspace{0pt}
\end{table}

\noindent \textbf{Out-of-sight completion.} We ablate the out-of-sight completion strategy. As shown in Table~\ref{tab:completion}, simple interpolation between boundary hand poses yields suboptimal results. Using hand poses only as temporal context for out-of-sight event completion already yields a noticeable performance gain. When further introducing head motion information, the accuracy improves substantially. This verifies our claim that out-of-sight completion can be performed in world space with head-hand coordination.

\begin{table}[t]
  \caption{Ablation of the out-of-sight completion. OW-p is the out-of-sight error of Table~\ref{tab:camera}, and W-MPJPE and WA-MPJPE are the world-space errors over the out-of-sight hands of Table~\ref{tab:world}. Best and second-best results are in bold black and green.}
  \label{tab:completion}
  \centering
  \footnotesize
  \setlength{\tabcolsep}{2pt}
  \setlength{\aboverulesep}{0pt}
  \setlength{\belowrulesep}{0pt}
  \renewcommand{\arraystretch}{1.1}
  \def\hs{\rule[-0.86ex]{0pt}{3.2ex}}%
  \def\hc#1{\raisebox{1.6ex}[0pt][0pt]{#1}}%
  \begin{tabularx}{\columnwidth}{l|Y|YY}
    \toprule
    \hs & \multicolumn{3}{c}{Out of sight} \\
    \cline{2-4}
    \multicolumn{1}{c|}{\hs\hc{Completion}} & OW-p$\downarrow$ & W-MPJPE$\downarrow$ & WA-MPJPE$\downarrow$ \\
    \midrule
    \noalign{\global\def\arraystretch{0.95}}
    \rule[-1.0ex]{0pt}{3.2ex}Interpolation & 336.94 & 528.93 & 301.24 \\
    Hand & \snd{161.01} & \snd{185.32} & \snd{114.64} \\
    \rowcolor{oursbg}\rule[-0.86ex]{0pt}{3.2ex}Hand + Head & \fst{131.07} & \fst{127.74} & \fst{91.83} \\
    \bottomrule
  \end{tabularx}
  \par\vspace{0pt}
\end{table}

\begin{figure}[t]
  \centering
  \includegraphics[width=\linewidth]{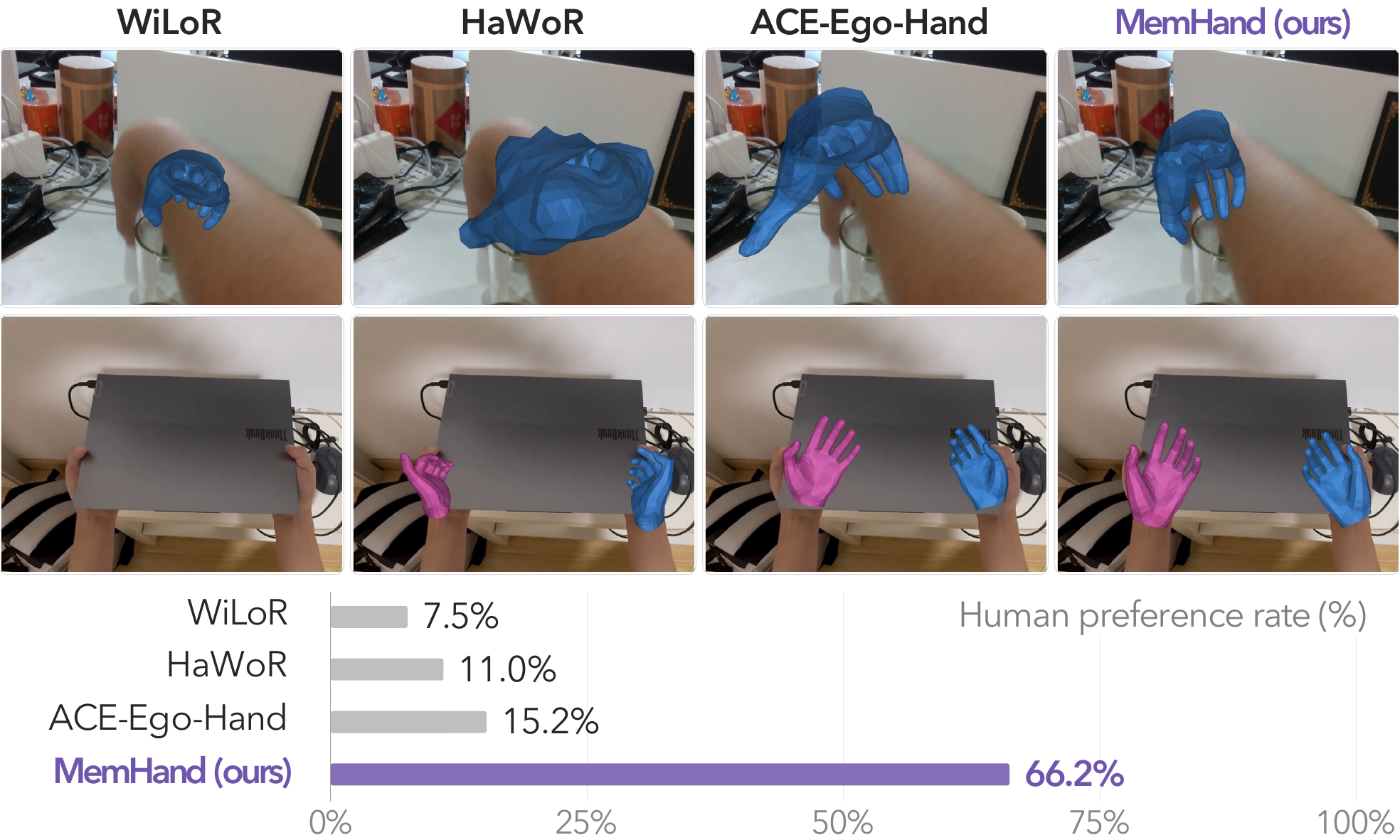}
  \caption{\textbf{Human preference study.} Top: qualitative hand motion reconstruction results on in-the-wild data. Bottom: the comparison of human preference rate over all choices.}
  \label{fig:preference}
\end{figure}

\subsection{Human Preference Study}
\label{sec:human}
Following~\cite{yang2026sam3dbody}, we conduct a human preference study on in-the-wild data to verify which hand reconstruction results align better with human perception. Specifically, we collect 400 egocentric videos using an iPhone and a RealSense D435i. We reconstruct hands on these videos with WiLoR, HaWoR, ACE-Ego-Hand, and MemHand, and render the results onto the original frames for visualization. A total of 53 in-house volunteers not involved in this project are shown anonymized reconstructions from the four methods in a randomly shuffled layout for each video, and asked to select the most visually plausible result. As shown in Fig.~\ref{fig:preference}, MemHand is preferred in 66.2\% of the comparisons. This could be attributed to MemHand exhibiting fewer false detections, identity flips, and temporal discontinuities under occlusion, yielding higher perceptual quality.

\subsection{Runtime Analysis}
\label{sec:runtime}
We further evaluate the efficiency of MemHand on an A100 GPU over the ARCTIC test sequences. Camera-space hand reconstruction alone runs at 26.1\,FPS, with the frozen V-JEPA encoder accounting for 57\% of the cost. For world-space reconstruction, it runs concurrently with camera tracking and depth estimation on separate CUDA streams of the same GPU, yielding an end-to-end throughput of 13.7\,FPS. As shown in Table~\ref{tab:runtime}, MemHand runs $1.6\times$ as fast as WiLoR-SLAM, $3.9\times$ as fast as HaWoR, and $12.5\times$ as fast as the optimization-based Dyn-HaMR, while being the most accurate. However, since the hands in a frame are decoded once the fixed window or event covering it closes, MemHand introduces a decoding lag, with 50\% of the reconstructed hands on both ARCTIC and HOT3D decoded within 101 frames and 95\% within 193 and 145 frames, respectively.

\begin{table}[t]
  \caption{Comparison of throughput and accuracy with world-space baselines on ARCTIC. W-MPJPE is the world-space error over all the hands of Table~\ref{tab:world}. Best results are in bold black.}
  \label{tab:runtime}
  \centering
  \footnotesize
  \setlength{\tabcolsep}{2pt}
  \setlength{\aboverulesep}{0pt}
  \setlength{\belowrulesep}{0pt}
  \renewcommand{\arraystretch}{1.1}
  \begin{tabularx}{\columnwidth}{l|Y|Y}
    \toprule
    \multicolumn{1}{c|}{\rule[-1.0ex]{0pt}{3.2ex}Method} & FPS$\uparrow$ & W-MPJPE$\downarrow$ \\
    \midrule
    \noalign{\global\def\arraystretch{0.95}}
    \rule[-1.0ex]{0pt}{3.2ex}WiLoR-SLAM~\cite{potamias2025wilor,teed2021droid,hu2024metric3dv2} & 8.5 & 130.55 \\
    Dyn-HaMR~\cite{yu2025dynhamr} & 1.1 & 184.12 \\
    HaWoR~\cite{zhang2025hawor} & 3.5 & 157.21 \\
    \rowcolor{oursbg}\rule[-0.86ex]{0pt}{3.2ex}MemHand (ours) & \fst{13.7} & \fst{41.46} \\
    \bottomrule
  \end{tabularx}
  \par\vspace{0pt}
\end{table}

\section{Conclusion}
\label{sec:conclusion}

In this paper, we propose a novel world-space hand motion estimation framework MemHand. It provides rich temporal context for occluded and out-of-sight segments via a devised short-term memory, and maintains accurate scale estimation through long-term memory. Experimental results validate that MemHand outperforms existing baselines on various metrics. Although the entire MemHand system achieves approximately 14 FPS on a single A100, it is still not real-time, considering the latency introduced by whole-event decoding. In the future, we plan to explore more efficient methods of temporal context utilization.


\bibliographystyle{ieeetran}

\footnotesize{
\bibliography{glorified, new}}

\end{document}

%% file: stachnisslab-latex.tex
\usepackage{graphics}           
\usepackage{times}              
\usepackage{amsmath}            
\usepackage{amssymb}            
\usepackage{graphicx}
\usepackage{algorithm}
\usepackage[noend]{algpseudocode}
\usepackage{booktabs}
\usepackage{color}
\definecolor{instructioncolor}{rgb}{.5,.5,.5}

\usepackage[font=small]{caption}

\def\eqref#1{Eq.~(\ref{#1})}

\makeatletter
\usepackage{xspace}
\DeclareRobustCommand\onedot{\futurelet\@let@token\@onedot}
\def\@onedot{\ifx\@let@token.\else.\null\fi\xspace}

\makeatother

\usepackage{array}
\newcolumntype{L}[1]{>{\raggedright\let\newline\\\arraybackslash\hspace{0pt}}m{#1}}
\newcolumntype{C}[1]{>{\centering\let\newline\\\arraybackslash\hspace{0pt}}m{#1}}
\newcolumntype{R}[1]{>{\raggedleft\let\newline\\\arraybackslash\hspace{0pt}}m{#1}}